\documentclass[10pt,twocolumn,letterpaper]{article}

\usepackage{cvpr}              

\usepackage[accsupp]{axessibility} 

\definecolor{cvprblue}{rgb}{0.21,0.49,0.74}
\usepackage[pagebackref,breaklinks,colorlinks,allcolors=cvprblue]{hyperref}

\def\paperID{42} 
\def\confName{CVPR}
\def\confYear{2026}

\title{Object-Level Explanations for Image Geolocation Models: a GeoGuessr use-case}

\author{Emilie Durrieu\\
ENAC, University of Toulouse\\
{\tt\small emilie.durrieu@utoulouse.fr}
\and
Christophe Hurter\\
ENAC, University of Toulouse\\
{\tt\small christohe.hurter@enac.fr}
\and
Philippe Muller\\
IRIT, University of Toulouse, ANITI\\
{\tt\small philippe.muller@irit.fr}
\and
Victor Boutin\\
CNRS\\
{\tt\small victor.boutin@cnrs.fr}
}

\begin{document}
\maketitle
\begin{abstract}
When humans play geolocation games such as GeoGuessr, they rely on concrete visual cues, such as road markings, vegetation, or architectural details, to infer where an image was captured. Whether image geolocation models rely on similar object-level evidence remains difficult to determine, as attribution methods like Grad-CAM typically highlight diffuse regions rather than coherent visual entities, making it difficult to link model predictions to specific objects or perceptible patterns.
In this work, we propose an object-centric analysis pipeline to investigate the visual evidence used by geolocation models. Starting from attribution maps, we extract salient regions and segment them into object-like elements. We evaluate their predictive relevance through deletion and insertion tests, comparing attribution-guided crops to randomly selected regions with similar coverage. Experiments on a three-country benchmark show that attribution-guided crops consistently retain more information for the model's prediction than random crops. These results suggest that attribution maps can  be decomposed into interpretable, perceptible elements, providing a step toward object-level analysis of geolocation models.
\end{abstract}    
\section{Introduction}
\label{sec:intro}

Image-based geolocation aims to predict where an image was captured using only its visual content. Beyond practical applications such as location verification or geographic indexing, this task has gained public attention through games such as \textit{Geoguessr} \footnote{\url{https://www.geoguessr.com}} and \textit{WorldGuessr}\footnote{\url{https://www.worldguessr.com}}. When humans solve these tasks, they typically rely on recognizable objects and structures, such as road signs, vegetation, vehicles, and architectural elements.

Modern Convolutional Neural Networks (CNNs) achieve strong performance by framing geolocation as a large-scale classification problem~\cite{weyland_planet_2016}. However, understanding which visual evidence drives these predictions remains difficult, limiting our ability to interpret, trust, and analyze geolocation models at a meaningful, concept-level granularity. Moreover, attribution maps such as those obtained with GradCAM~\cite{selvaraju_grad-cam_2017} often highlight broad or overlapping regions that do not correspond to discrete, perceptible objects, making it difficult to link model decisions to interpretable visual patterns.

\begin{figure*}[t]
    \centering
    \includegraphics[width=1\textwidth]{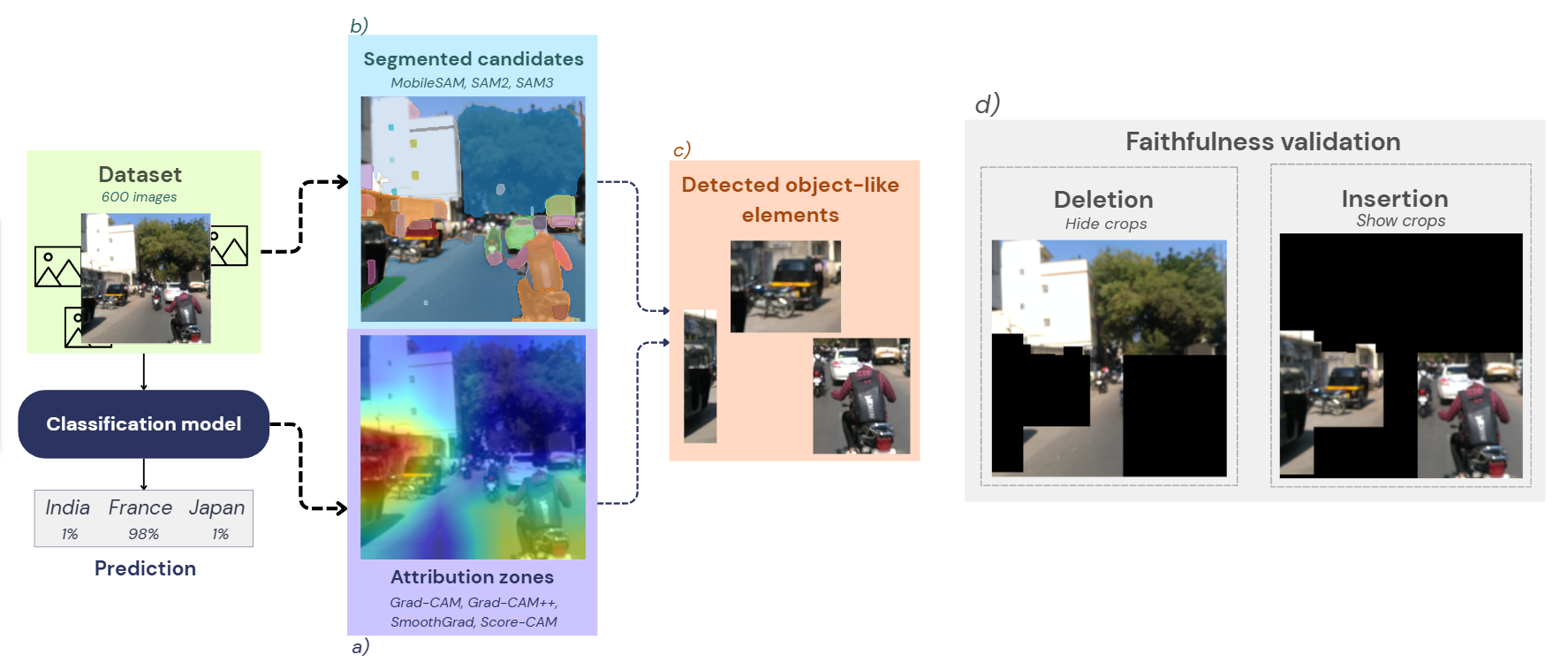}
    \caption{
    Overview of the proposed object-centric pipeline: (a) Saliency maps are extracted from a trained classifier using GradCAM++. (b) Images are segmented into candidate regions. (c) Each segment is scored based on overlap with attribution maps to identify relevant object-like elements. (d) Faithfulness validation is performed with a insertion/deletion test to assess their predictive relevance.
    }
    \label{fig:pipeline}
\end{figure*}

To address this, we propose an object-centric analysis pipeline for geolocation interpretability. Attribution maps produced from a trained classifier are first thresholded to identify salient regions. These regions are then segmented into object-like elements using a segmentation model, producing discrete visual units that can be analyzed individually. Finally, we evaluate whether these extracted elements preserve predictive information through deletion and insertion-based faithfulness tests (see Figure~\ref{fig:pipeline}).

Our experiments on a three-country benchmark show that attribution-guided object-like elements retain more predictive information than randomly selected regions of similar coverage. These results suggest that attribution maps contain localized evidence that can be meaningfully decomposed into perceptible visual regions, opening new opportunities for object-level interpretability and concept-based explanations in image geolocation models.

\section{Related Work}
\label{sec:formatting}

\subsection{Geolocation Models}
Automatic image geolocation aims to predict the location where an image was taken, using either global descriptors or deep learning features. Early methods such as IM2GPS~\cite{hays_im2gps_2008} relied on hand-crafted features and nearest-neighbor retrieval over large image databases. More recent approaches employ deep convolutional neural networks to formulate geolocation as a classification problem: PlaNet~\cite{weyland_planet_2016} divides the world into discrete cells and predicts the most likely location, while IM2GPS Revisited~\cite{vo_revisiting_2017} and PIGEON~\cite{haas_pigeon_2024} improve accuracy using CNN features and hierarchical modeling. These methods achieve impressive performance on both world-scale and street-level datasets, but they generally focus on prediction accuracy rather than understanding the visual cues that drive their decisions.

\subsection{Explainable AI in vision}
Attribution and visualization methods in XAI aim to reveal which parts of an image influence a model's prediction. Gradient-based methods such as GradCAM~\cite{selvaraju_grad-cam_2017} and Integrated Gradients~\cite{sundararajan_integratedgradients_2017} generate class-specific attention maps from convolutional features, while perturbation-based methods like LIME~\cite{ribeiro_lime_2016} and RISE~\cite{petsiuk_rise_2018} estimate importance by observing output changes under input modifications. Gradient-based approaches are particularly convenient for our study as they provide smooth and spatially coherent maps that can be efficiently computed and combined with segmentation to isolate object-like regions.
Applying XAI to geolocation models is relatively unexplored at the object level. Shi et al.~\cite{shi_deep_2019} visualize city-level CNN predictions to highlight relevant regions, showing that urban structures and landmarks drive the model's decisions. Our work builds on this idea but focuses on object-level interpretability, extracting object-like elements from attribution-guided regions to study their relevance.

\subsection{Concept-based XAI}
Concept-based explanations provide a more structured view of model reasoning by linking predictions to human-interpretable concepts. Concept Bottleneck Models~\cite{koh_cbm_2020} train networks to first predict predefined concepts before outputting a final label, ensuring interpretability by design. Automated Concept-based Explanation (ACE)~\cite{ghorbani_ace_2019} extracts visual concepts from images using segmentation and clustering, then evaluates their relevance to model predictions. In contrast, our approach focuses on the geolocation context and extracts object-like elements directly from attribution maps, without relying on predefined concepts or modifying the model. This enables a post-hoc, object-level analysis of the visual cues driving geolocation predictions.

\section{Methodology}
\label{sec:method}

We investigate whether visual regions highlighted by attribution methods capture predictive information in geolocation models. Our approach consists of three main steps illustrated in Figure~\ref{fig:pipeline}: (1) Train a CNN classifier for geolocation. (2) Extract attribution-guided regions from model predictions. (3) Decompose these regions into object-like elements and evaluate their predictive relevance.

\subsection{Problem Setup}

Let $f$ be a Convolutional Neural Network (CNN) trained to predict the country of origin among $N$ possible countries. Given an input image $x$ representing a ground-level geolocation image, the model outputs a logit vector $f(x)$ in $\mathbb{R}^N$, which is then converted into a probability distribution via softmax.
Our goal is to identify the visual cues present in $x$ that contribute to the model's predictions.

\subsection{Obtaining visual explanations}

To localize the most predictive image regions, we apply feature attribution methods to our model $f$. For a given input image and predicted class, these methods produce an attention map highlighting pixels that strongly influence the output. We threshold the map to retain the most salient pixels by selecting the top-$p$ percentile of pixels based on attribution values, with $p$ chosen empirically by evaluating the resulting coverage and faithfulness on a validation set. This strategy balances keeping highly informative pixels while avoiding overly large or diffuse regions. We refer to the obtained salient maps as \emph{attribution-guided regions}.

\subsection{Extracting object-like elements}

Within each attribution-guided region, we extract discrete visual elements using a segmentation model. We define an \emph{object-like element} as a perceptible visual unit, such as a car or a street sign. These elements are not assigned semantic labels, and their shapes depend on the segmentation output.
Each candidate segment is assigned a relevance score based on three factors:
\begin{enumerate}
    \item \textbf{Overlap with high-saliency regions:} the fraction of the segment that overlaps pixels above a threshold in the attribution map.
    \item \textbf{Average importance:} the mean attribution value within the segment.
    \item \textbf{Central importance:} the attribution value at the segment’s geometric center.
\end{enumerate}
These factors are then combined using the geometric mean.

Segments below threshold $s_{\min}$ are discarded, and the remaining segments are ranked by score.
To avoid redundancy, overlapping segments are filtered using a containment-based IoU criterion, which ensures small details are preserved while near-duplicate segments are removed. Finally, each segment is converted into a rectangular bounding box and slightly padded to preserve contextual information. These rectangular zones constitute the final set of object-like elements used for evaluation.
\section{Our extraction method}
\label{sec:experiments}


\textbf{Dataset}: We use images from the dataset OSV-5M~\cite{astruc_osv5m_2024} containing over 5 million street-level images from 225 different countries collected via the Mapillary API. For our study, we select images from three countries: France, India, and Japan, and retrieve the original high-resolution images through the API. The data is split into 100k images (33,333 per country) for training and 600 images (200 per country) for pipeline evaluation.
All images are resized to $224 \times 224$ pixels to match the input resolution of the classifier. Pixel values are normalized using ImageNet statistics. During training, standard data augmentation is applied, including random horizontal flipping and color jitter, while evaluation images are processed without augmentation.

\noindent\textbf{Classification Model}: We finetune a ResNet50~\cite{he_resnet50_2016} pretrained on ImageNet~\cite{deng_imagenet_2009} for country classification. The network is trained with Cross-Entropy loss with label smoothing of 0.1, using the AdamW optimizer. The learning rate was set to 3e-4, with a weight decay of 0.02. The model was trained for 300 epochs with early stopping based on validation loss (patience = 30). The model reached an accuracy of 88\% on the training set and 87.42\% on the test set, as well as 86.3\% on the 600 pipeline images.

\noindent\textbf{Attention regions}: To identify regions that influence the model's predictions, we apply several attribution methods: Grad-CAM, Grad-CAM++, SmoothGrad, and Score-CAM.
GradCAM~\cite{selvaraju_grad-cam_2017} produces class-specific localization maps by weighting convolutional feature maps using gradients, while GradCAM++~\cite{chattopadhay_grad-campp_2018} refines this process by using pixel-wise and higher-order gradient weighting, thereby improving multi-instance localization. SmoothGrad~\cite{smilkov_smoothgrad_2017} reduces noise by averaging gradients over multiple perturbed inputs, and Score-CAM~\cite{wang_score-cam_2020} generates class activation maps in a gradient-free manner using forward-pass class scores.

\noindent\textbf{Object extraction}: Within attribution-guided regions, we extract perceptible object-like elements using segmentation models. We experiment with three Segment Anything Model (SAM) variants: lightweight \textit{MobileSAM}~\cite{liu_mobilesam_2023}, \textit{SAM2}~\cite{ravi_sam2_2024}, and the recent \textit{SAM3}~\cite{carion_sam3_2025} which extends SAM2 with concept-guided prompting. For SAM3, the employed concepts are extracted from a Geoguessr guidebook using an LLM, resulting in roughly 200 candidate concepts associated with geolocation.
\section{Evaluation}
\label{sec:evaluation}

The goal of this evaluation is to determine whether the object-like elements extracted by our pipeline correspond to visual regions that are genuinely important for the geolocation model's prediction. In particular, we investigate whether these extracted regions retain predictive information by performing two faithfulness tests commonly used in explainable AI:

\begin{enumerate}
    \item \emph{Deletion}: Extracted crops are occluded in the image, and we measure the drop in classification accuracy. A larger drop indicates that these removed crops contained important information for the model's prediction.
    \item \emph{Insertion}: Only the extracted crops are preserved while the rest of the image is masked. The model predicts from this reduced input, and the retained classification accuracy quantifies how much predictive information is contained within the object-like elements.
\end{enumerate}

\begin{figure}[t]
\centering
\includegraphics[width=0.9\linewidth]{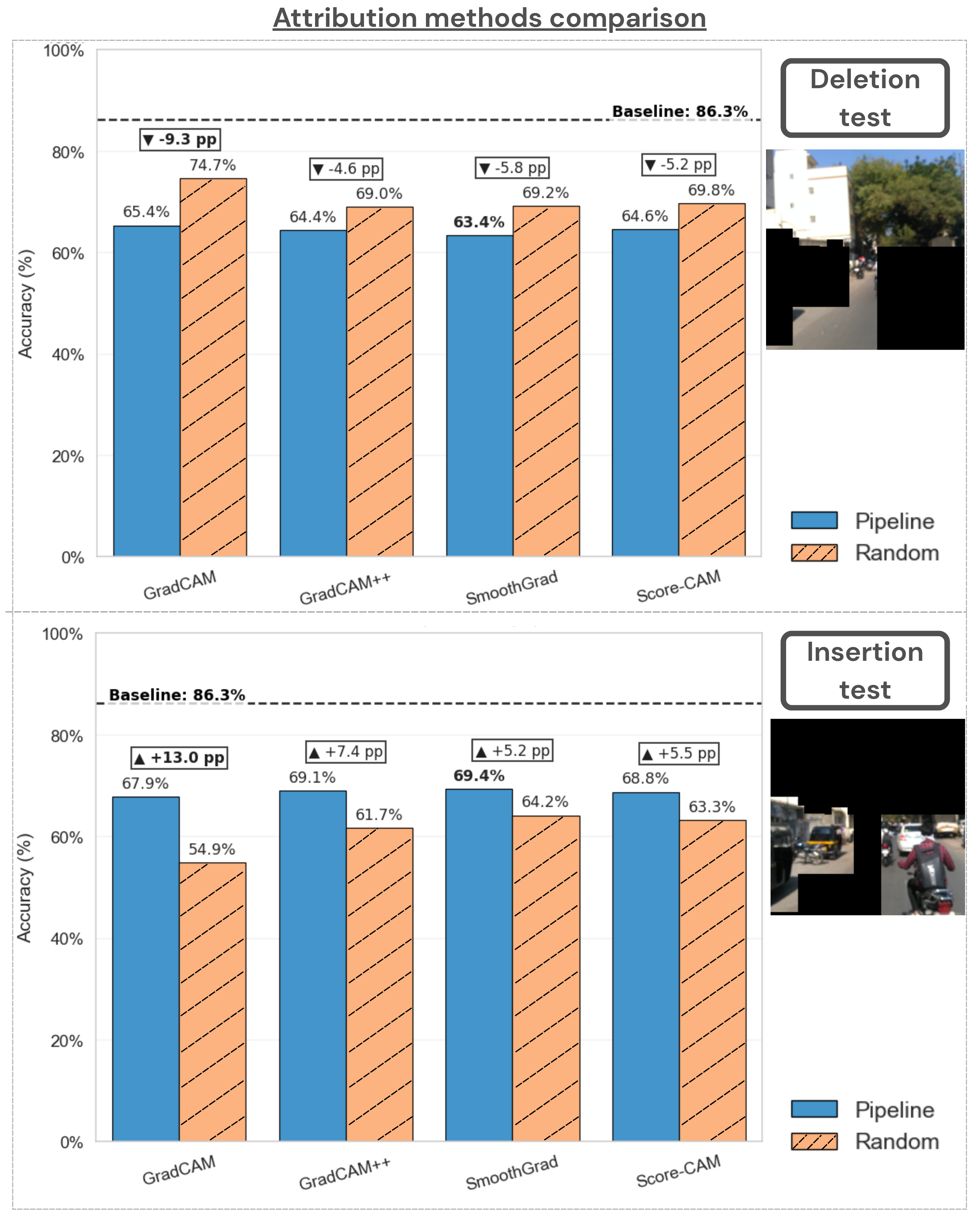}
\caption{
Comparison of attribution methods (GradCAM, GradCAM++, SmoothGrad, Score-CAM) using MobileSAM as the segmentation model.
}
\label{fig:attr_validation}
\end{figure}

\begin{figure}[t]
\centering
    \includegraphics[width=0.9\linewidth]{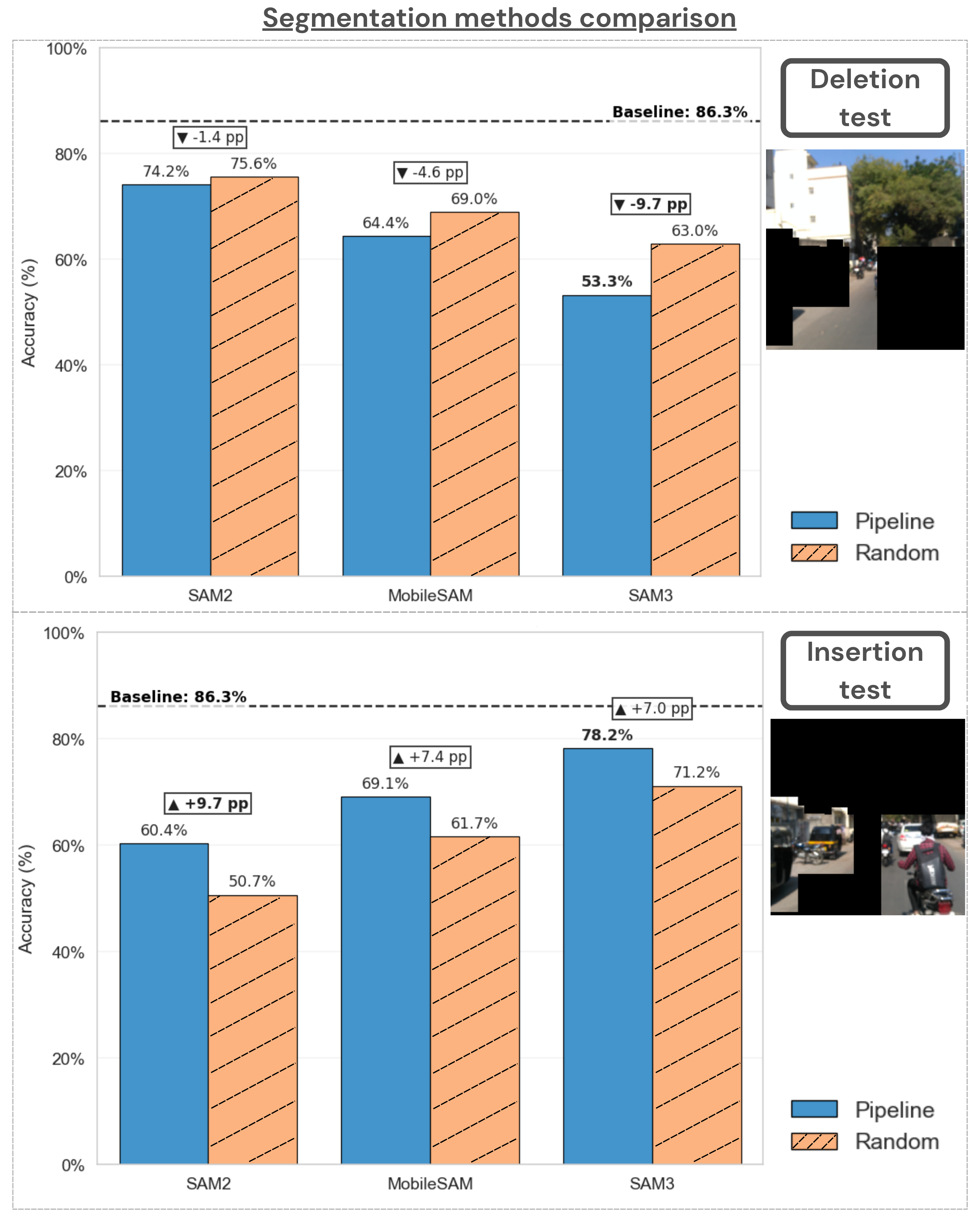}
\caption{
Comparison of segmentation methods (MobileSAM, SAM2, SAM3) using GradCAM++ as the attribution method.
}
\label{fig:seg_validation}
\end{figure}

For comparison, we generate random crops with similar coverage and size constraints. The procedure is repeated ten times per image, and the results are averaged to reduce variance. This baseline allows us to determine whether attribution-guided extraction identifies more informative regions than random selection.

We further analyze how different attribution methods (Figure~\ref{fig:attr_validation}) and segmentation strategies (Figure~\ref{fig:seg_validation}) influence the effectiveness of the extracted regions.

Across all tested configurations, attribution-guided crops consistently outperform random crops. In the deletion test, occluding these regions causes a larger drop in accuracy, indicating that they correspond to areas that are important for the model's predictions. In the insertion test, retaining only these regions preserves significantly more predictive information than random crops of similar coverage. These results suggest that attribution maps contain structured predictive signals that can be decomposed into localized visual elements without losing task-relevant information.

Interestingly, segmentation quality appears to influence the results. As shown in Figure~\ref{fig:seg_validation}, SAM2 performs poorly in the Deletion test, which may be explained by the relatively small number of segments it generates. This results in lower spatial coverage of the image, limiting the amount of information removed when the crops are ablated.
In contrast, SAM3 produces a significantly larger number of segments than the other methods and achieves stronger results in both ablation settings. This suggests that higher segmentation coverage enables the pipeline to capture a broader set of visual cues contributing to the model's predictions.

\begin{figure}[t]
\centering
    \includegraphics[width=1\linewidth]{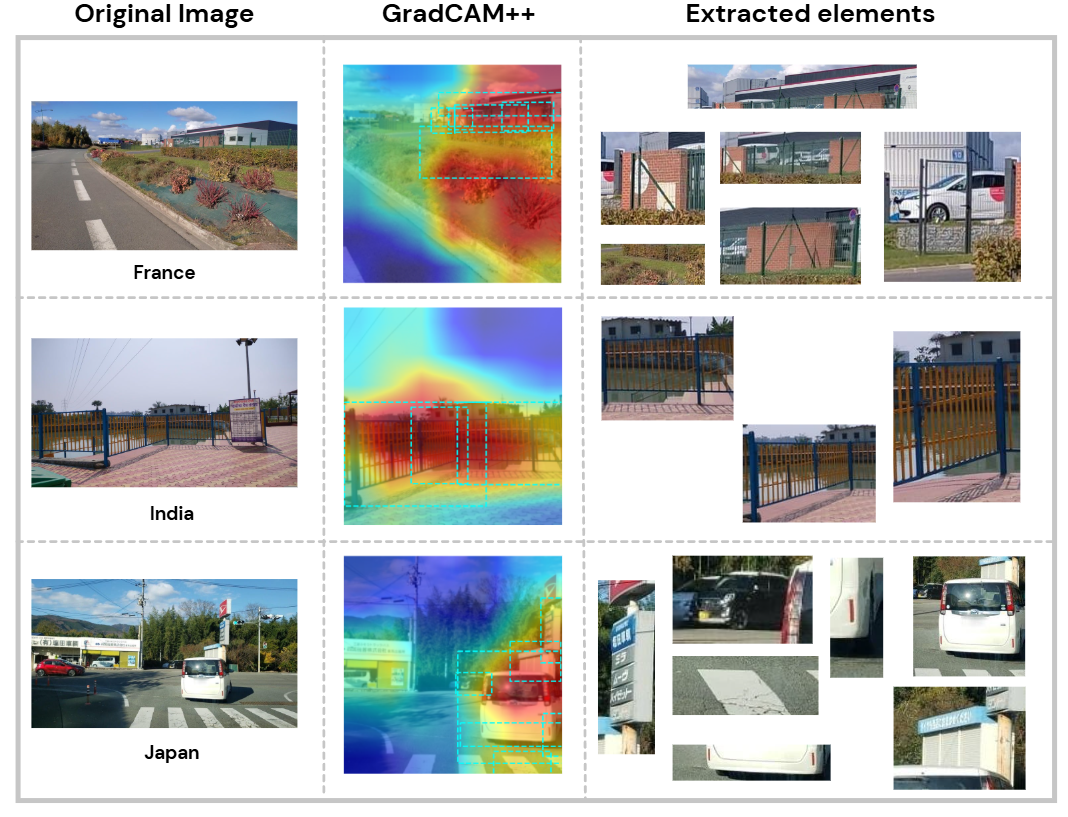}
\caption{
Attribution-guided object-like elements extracted from geolocation images. The method identifies diverse visual cues, including vehicles, road markings, and barriers, although some regions correspond to partial or overlapping structures.
}
\label{fig:examples}
\end{figure}

Qualitative inspection of the extracted object-like elements reveals a diverse set of visual patterns, including vehicles, walls, street signs, and road markings (see Figure~\ref{fig:examples}). This diversity suggests that the pipeline captures a range of perceptible cues that can contribute to geolocation predictions. At the same time, some elements overlap or correspond only to parts of larger structures, while others remain amorphous, reflecting limitations of the segmentation step.

\section{Conclusion}




We present a post-hoc, object-centric framework for interpreting image geolocation models by transforming attribution maps into discrete, object-like elements that can be evaluated directly. By combining existing attribution and segmentation methods as well as deletion/insertion faithfulness tests, our approach bridges diffuse pixel-level explanations and localized visual evidence without requiring predefined concepts or modifications to the underlying classifier. On a three-country benchmark, attribution-guided crops consistently outperform random regions with comparable coverage, indicating that the extracted elements retain meaningful predictive information and that predictions can be traced to perceptible visual cues. Our results also suggest that segmentation quality plays an important role in this analysis, as broader and more appropriate region coverage leads to stronger faithfulness outcomes.

At the same time, the extracted elements remain approximate and may be over-segmented, and our evaluation is limited to a three-country setting without additional baselines. Future work includes extending to larger, fine-grained benchmarks, explore alternative attribution and region-selection strategies, and incorporate human evaluation to assess whether the visual evidence identified by the model aligns with human reasoning in geolocation tasks such as GeoGuessr.





{
    \small
    \bibliographystyle{ieeenat_fullname}
    \bibliography{main}

@inproceedings{koh_cbm_2020,
  title = {Concept {{Bottleneck Models}}},
  booktitle = {Proceedings of the 37th {{International Conference}} on {{Machine Learning}}},
  author = {Koh, Pang Wei and Nguyen, Thao and Tang, Yew Siang and Mussmann, Stephen and Pierson, Emma and Kim, Been and Liang, Percy},
  year = 2020,
  month = nov,
  pages = {5338--5348},
  publisher = {PMLR},
  issn = {2640-3498},
  urldate = {2025-10-08},
  langid = {english},
}

@misc{shi_deep_2019,
  title = {Deep {{Visual City Recognition Visualization}}},
  author = {Shi, Xiangwei and Khademi, Seyran and van Gemert, Jan},
  year = 2019,
  month = may,
  number = {arXiv:1905.01932},
  eprint = {1905.01932},
  primaryclass = {cs},
  publisher = {arXiv},
  doi = {10.48550/arXiv.1905.01932},
  urldate = {2025-10-09},
  archiveprefix = {arXiv},
}

@inproceedings{hays_im2gps_2008,
  title = {{{IM2GPS}}: Estimating Geographic Information from a Single Image},
  shorttitle = {{{IM2GPS}}},
  booktitle = {2008 {{IEEE Conference}} on {{Computer Vision}} and {{Pattern Recognition}}},
  author = {Hays, James and Efros, Alexei A.},
  year = 2008,
  month = jun,
  pages = {1--8},
  issn = {1063-6919},
  doi = {10.1109/CVPR.2008.4587784},
  urldate = {2025-10-09},
}

@inproceedings{vo_revisiting_2017,
  title = {Revisiting {{IM2GPS}} in the {{Deep Learning Era}}},
  booktitle = {2017 {{IEEE International Conference}} on {{Computer Vision}} ({{ICCV}})},
  author = {Vo, Nam and Jacobs, Nathan and Hays, James},
  year = 2017,
  month = oct,
  pages = {2640--2649},
  issn = {2380-7504},
  doi = {10.1109/ICCV.2017.286},
  urldate = {2025-10-09},
}

@inproceedings{astruc_osv5m_2024,
  title = {{{OpenStreetView-5M}}: {{The Many Roads}} to {{Global Visual Geolocation}}},
  shorttitle = {{{OpenStreetView-5M}}},
  booktitle = {2024 {{IEEE}}/{{CVF Conference}} on {{Computer Vision}} and {{Pattern Recognition}} ({{CVPR}})},
  author = {Astruc, Guillaume and Dufour, Nicolas and Siglidis, Ioannis and Aronssohn, Constantin and Bouia, Nacim and Fu, Stephanie and Loiseau, Romain and Nguyen, Van Nguyen and Raude, Charles and Vincent, Elliot and Xu, Lintao and Zhou, Hongyu and Landrieu, Loic},
  year = 2024,
  month = jun,
  pages = {21967--21977},
  issn = {2575-7075},
  doi = {10.1109/CVPR52733.2024.02074},
  urldate = {2025-10-09},
}

@inproceedings{haas_pigeon_2024,
  title = {{{PIGEON}}: {{Predicting Image Geolocations}}},
  shorttitle = {{{PIGEON}}},
  booktitle = {2024 {{IEEE}}/{{CVF Conference}} on {{Computer Vision}} and {{Pattern Recognition}} ({{CVPR}})},
  author = {Haas, Lukas and Skreta, Michal and Alberti, Silas and Finn, Chelsea},
  year = 2024,
  month = jun,
  pages = {12893--12902},
  issn = {2575-7075},
  doi = {10.1109/CVPR52733.2024.01225},
  urldate = {2025-10-09},
}

@inproceedings{weyland_planet_2016,
  title = {{{PlaNet}} - {{Photo Geolocation}} with {{Convolutional Neural Networks}}},
  booktitle = {Computer {{Vision}} -- {{ECCV}} 2016},
  author = {Weyand, Tobias and Kostrikov, Ilya and Philbin, James},
  editor = {Leibe, Bastian and Matas, Jiri and Sebe, Nicu and Welling, Max},
  year = 2016,
  pages = {37--55},
  publisher = {Springer International Publishing},
  address = {Cham},
  doi = {10.1007/978-3-319-46484-8_3},
  isbn = {978-3-319-46484-8},
  langid = {english},
}

@inproceedings{ribeiro_lime_2016,
  title = {``{{Why Should I Trust You}}?'': {{Explaining}} the {{Predictions}} of {{Any Classifier}}},
  shorttitle = {``{{Why Should I Trust You}}?},
  booktitle = {Proceedings of the 2016 {{Conference}} of the {{North American Chapter}} of the {{Association}} for {{Computational Linguistics}}: {{Demonstrations}}},
  author = {Ribeiro, Marco and Singh, Sameer and Guestrin, Carlos},
  editor = {DeNero, John and Finlayson, Mark and Reddy, Sravana},
  year = 2016,
  month = jun,
  pages = {97--101},
  publisher = {Association for Computational Linguistics},
  address = {San Diego, California},
  doi = {10.18653/v1/N16-3020},
  urldate = {2025-10-09},
}

@inproceedings{selvaraju_grad-cam_2017,
  title = {Grad-{{CAM}}: {{Visual Explanations}} from {{Deep Networks}} via {{Gradient-Based Localization}}},
  shorttitle = {Grad-{{CAM}}},
  booktitle = {2017 {{IEEE International Conference}} on {{Computer Vision}} ({{ICCV}})},
  author = {Selvaraju, Ramprasaath R. and Cogswell, Michael and Das, Abhishek and Vedantam, Ramakrishna and Parikh, Devi and Batra, Dhruv},
  year = 2017,
  month = oct,
  pages = {618--626},
  issn = {2380-7504},
  doi = {10.1109/ICCV.2017.74},
  urldate = {2025-10-09},
}

@inproceedings{ghorbani_ace_2019,
  title = {Towards {{Automatic Concept-based Explanations}}},
  booktitle = {Advances in {{Neural Information Processing Systems}}},
  author = {Ghorbani, Amirata and Wexler, James and Zou, James Y and Kim, Been},
  year = 2019,
  volume = {32},
  publisher = {Curran Associates, Inc.},
  urldate = {2025-10-10},
}

@misc{carion_sam3_2025,
  title = {{{SAM}} 3: {{Segment Anything}} with {{Concepts}}},
  shorttitle = {{{SAM}} 3},
  author = {Carion, Nicolas and Gustafson, Laura and Hu, Yuan-Ting and Debnath, Shoubhik and Hu, Ronghang and Suris, Didac and Ryali, Chaitanya and Alwala, Kalyan Vasudev and Khedr, Haitham and Huang, Andrew and Lei, Jie and Ma, Tengyu and Guo, Baishan and Kalla, Arpit and Marks, Markus and Greer, Joseph and Wang, Meng and Sun, Peize and R{\"a}dle, Roman and Afouras, Triantafyllos and Mavroudi, Effrosyni and Xu, Katherine and Wu, Tsung-Han and Zhou, Yu and Momeni, Liliane and Hazra, Rishi and Ding, Shuangrui and Vaze, Sagar and Porcher, Francois and Li, Feng and Li, Siyuan and Kamath, Aishwarya and Cheng, Ho Kei and Doll{\'a}r, Piotr and Ravi, Nikhila and Saenko, Kate and Zhang, Pengchuan and Feichtenhofer, Christoph},
  year = 2025,
  month = nov,
  number = {arXiv:2511.16719},
  eprint = {2511.16719},
  primaryclass = {cs},
  publisher = {arXiv},
  doi = {10.48550/arXiv.2511.16719},
  urldate = {2026-02-17},
  archiveprefix = {arXiv},
}

@inproceedings{chattopadhay_grad-campp_2018,
  title = {Grad-{{CAM}}++: {{Generalized Gradient-Based Visual Explanations}} for {{Deep Convolutional Networks}}},
  shorttitle = {Grad-{{CAM}}++},
  booktitle = {2018 {{IEEE Winter Conference}} on {{Applications}} of {{Computer Vision}} ({{WACV}})},
  author = {Chattopadhay, Aditya and Sarkar, Anirban and Howlader, Prantik and Balasubramanian, Vineeth N},
  year = 2018,
  month = mar,
  pages = {839--847},
  doi = {10.1109/WACV.2018.00097},
  urldate = {2026-03-05},
}

@article{liu_mobilesam_2023,
  title = {{{MobileSAM-Track}}: {{Lightweight One-Shot Tracking}} and {{Segmentation}} of {{Small Objects}} on {{Edge Devices}}},
  shorttitle = {{{MobileSAM-Track}}},
  author = {Liu, Yehui and Zhao, Yuliang and Zhang, Xinyue and Wang, Xiaoai and Lian, Chao and Li, Jian and Shan, Peng and Fu, Changzeng and Lyu, Xiaoyong and Li, Lianjiang and Fu, Qiang and Li, Wen Jung},
  year = 2023,
  month = dec,
  journal = {Remote Sensing},
  volume = {15},
  number = {24},
  publisher = {Multidisciplinary Digital Publishing Institute},
  issn = {2072-4292},
  doi = {10.3390/rs15245665},
  urldate = {2026-03-05},
  copyright = {http://creativecommons.org/licenses/by/3.0/},
  langid = {english},
}

@misc{petsiuk_rise_2018,
  title = {{{RISE}}: {{Randomized Input Sampling}} for {{Explanation}} of {{Black-box Models}}},
  shorttitle = {{{RISE}}},
  author = {Petsiuk, Vitali and Das, Abir and Saenko, Kate},
  year = 2018,
  month = sep,
  number = {arXiv:1806.07421},
  eprint = {1806.07421},
  primaryclass = {cs},
  publisher = {arXiv},
  doi = {10.48550/arXiv.1806.07421},
  urldate = {2026-03-06},
  archiveprefix = {arXiv},
}

@inproceedings{he_resnet50_2016,
  title = {Deep {{Residual Learning}} for {{Image Recognition}}},
  booktitle = {Proceedings of the {{IEEE Conference}} on {{Computer Vision}} and {{Pattern Recognition}}},
  author = {He, Kaiming and Zhang, Xiangyu and Ren, Shaoqing and Sun, Jian},
  year = 2016,
  pages = {770--778},
  urldate = {2026-03-06},
}

@inproceedings{deng_imagenet_2009,
  title = {{{ImageNet}}: {{A}} Large-Scale Hierarchical Image Database},
  shorttitle = {{{ImageNet}}},
  booktitle = {2009 {{IEEE Conference}} on {{Computer Vision}} and {{Pattern Recognition}}},
  author = {Deng, Jia and Dong, Wei and Socher, Richard and Li, Li-Jia and Li, Kai and {Fei-Fei}, Li},
  year = 2009,
  month = jun,
  pages = {248--255},
  issn = {1063-6919},
  doi = {10.1109/CVPR.2009.5206848},
  urldate = {2026-03-06},
}

@inproceedings{sundararajan_integratedgradients_2017,
  title = {Axiomatic {{Attribution}} for {{Deep Networks}}},
  booktitle = {Proceedings of the 34th {{International Conference}} on {{Machine Learning}}},
  author = {Sundararajan, Mukund and Taly, Ankur and Yan, Qiqi},
  year = 2017,
  month = jul,
  pages = {3319--3328},
  publisher = {PMLR},
  issn = {2640-3498},
  urldate = {2026-03-06},
  langid = {english},
}

@inproceedings{wang_score-cam_2020,
  title = {Score-{{CAM}}: {{Score-Weighted Visual Explanations}} for {{Convolutional Neural Networks}}},
  shorttitle = {Score-{{CAM}}},
  booktitle = {Proceedings of the {{IEEE}}/{{CVF Conference}} on {{Computer Vision}} and {{Pattern Recognition Workshops}}},
  author = {Wang, Haofan and Wang, Zifan and Du, Mengnan and Yang, Fan and Zhang, Zijian and Ding, Sirui and Mardziel, Piotr and Hu, Xia},
  year = 2020,
  pages = {24--25},
  urldate = {2026-03-06},
}

@misc{smilkov_smoothgrad_2017,
  title = {{{SmoothGrad}}: Removing Noise by Adding Noise},
  shorttitle = {{{SmoothGrad}}},
  author = {Smilkov, Daniel and Thorat, Nikhil and Kim, Been and Vi{\'e}gas, Fernanda and Wattenberg, Martin},
  year = 2017,
  month = jun,
  number = {arXiv:1706.03825},
  eprint = {1706.03825},
  primaryclass = {cs},
  publisher = {arXiv},
  doi = {10.48550/arXiv.1706.03825},
  urldate = {2026-03-06},
  archiveprefix = {arXiv},
}

@misc{ravi_sam2_2024,
  title = {{{SAM}} 2: {{Segment Anything}} in {{Images}} and {{Videos}}},
  shorttitle = {{{SAM}} 2},
  author = {Ravi, Nikhila and Gabeur, Valentin and Hu, Yuan-Ting and Hu, Ronghang and Ryali, Chaitanya and Ma, Tengyu and Khedr, Haitham and R{\"a}dle, Roman and Rolland, Chloe and Gustafson, Laura and Mintun, Eric and Pan, Junting and Alwala, Kalyan Vasudev and Carion, Nicolas and Wu, Chao-Yuan and Girshick, Ross and Doll{\'a}r, Piotr and Feichtenhofer, Christoph},
  year = 2024,
  month = oct,
  number = {arXiv:2408.00714},
  eprint = {2408.00714},
  primaryclass = {cs},
  publisher = {arXiv},
  doi = {10.48550/arXiv.2408.00714},
  urldate = {2026-03-06},
  archiveprefix = {arXiv},
}
}


\end{document}